\documentclass{article}
\usepackage{spconf,amsmath,graphicx}
\usepackage{amsfonts} 
\usepackage{booktabs}
\usepackage{url}

\title{INTERPRETABLE VISUAL QUESTION ANSWERING VIA REASONING SUPERVISION}

\name{Maria Parelli$^{\dagger \ddagger}$ \qquad Dimitrios Mallis$^{\dagger}$ \qquad Markos Diomataris$^{\dagger \ddagger}$\sthanks{Work was done while Markos Diomataris was with DeepLab.} \qquad Vassilis Pitsikalis$^{\dagger}$}
  
  \address{$^{\dagger}$ DeepLab, Athens, Greece\\
      $^{\ddagger}$ ETH Z{\"u}rich, Z{\"u}rich, Switzerland  
    \\
    {\small \url{maryparelli@gmail.com}, \url{mallidimi1@gmail.com}, \url{mdiomataris@student.ethz.ch}, \url{vpitsik@deeplab.ai}}}

\begin{document}
%
\maketitle
\begin{abstract}
Transformer-based architectures have recently demonstrated remarkable performance in the Visual Question Answering (VQA) task. However, such models are likely to disregard crucial visual cues and often rely on multimodal shortcuts and inherent biases of the language modality to predict the correct answer, a phenomenon commonly referred to as \textit{lack of visual grounding}. In this work, we alleviate this shortcoming through a novel architecture for visual question answering that leverages \textit{common sense reasoning as a supervisory signal}. Reasoning supervision takes the form of a textual justification of the correct answer, with such annotations being already available on large-scale Visual Common Sense Reasoning (VCR) datasets. The model's visual attention is guided toward important elements of the scene through a similarity loss that aligns the learned attention distributions guided by the question and the correct reasoning. We demonstrate both quantitatively and qualitatively that the proposed approach can boost the model's visual perception capability and lead to performance increase, without requiring training on explicit grounding annotations.
\end{abstract}
\begin{keywords}
Visual Question Answering, Visual Grounding, Interpretability, Attention Similarity
\end{keywords}
\section{Introduction}
\label{sec:intro}
Models for Visual Question Answering (VQA) provide answers to natural language questions about an image by perceiving both textual and image cues. VQA lies at the intersection of vision and language and has recently generated significant research interest. Existing methods aim to tackle the task via deep multi-layer transformer architectures, attending to linguistic and visual tokens \cite{oscar,VLBERT,uniter,vinvl,lxmert}.  However, despite their superior performance, attempts to diagnose these models' robustness and reasoning capability have revealed that they often rely on linguistic biases and shallow correlations to generate the correct answer \cite{mutant,robust}. The language modality has been proven a strong signal that is easy to exploit, causing the model to overlook visual information and rely on shallow patterns, such as correlations between words in the question \cite{matter}. It has been shown that the performance of recent models can clearly degrade under evaluation settings that penalize reliance on such spurious correlations~\cite{dja,shortcuts,swapmix}.

 \begin{figure}
  \centering
  \includegraphics[scale=0.28]{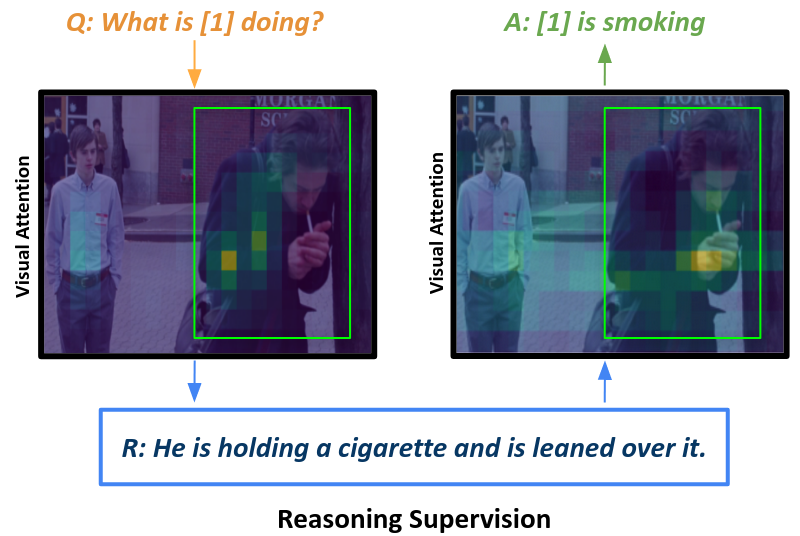}
  \caption{This work proposes a novel mechanism for leveraging \textit{common sense reasoning as a supervisory signal}. Our VQA model, guided by the correct reasoning \textsc{(R:[person1] is holding the cigarette)}, is able to attend to the appropriate image regions and accurately select the right answer \textsc{(A:[person1] is smoking)}.}
    \label{fig:intro}
\end{figure}
This tendency of recent models to reason about the correct answer without attending to the relevant image areas has been referred to as \textit{lack of visual grounding}~\cite{hint,impaired}. To alleviate this, a line of work explores techniques for training VQA models that are sensitive to the same image regions as human annotators, commonly by enforcing alignment with human attention maps~\cite{ hint,human}. While such methods can reduce reliance on language biases, they also require explicit grounding supervision that is rarely available. In this work, we explore an alternative approach towards attending to informative image regions, that does not require explicit grounding supervision, but leverages instead \textit{common sense reasoning} as a supervisory signal.

We take advantage of the fact that reasoning-level supervision in the form of textual justification of why an answer is true, is already available in large-scale Visual Commonsense Reasoning datasets like \cite{vcr}. For example, in Fig. \ref{fig:intro}, to answer the \textbf{question} \textit{`What is \textsc{[person1]} doing?'}, the \textbf{reasoning} \textit{'\textsc{[person1]} is holding a cigarette and is leaned over it'} can accurately guide a model's visual attention towards predicting the correct \textbf{answer}, \textit{`\textsc{[person1]} is smoking'}. The correct reasoning often contains details of the scene and references to objects and people relevant to the right answer. Our VQA model is trained to utilize reasoning supervision as a proxy signal to generate interpretable attention maps that guide visual attention toward informative image regions.

Our proposed framework processes question/answer pairs using a multilayer BERT~\cite{bert} transformer architecture. A separate visual attention stream is incorporated to generate two attention distributions, one conditioned on the question and the other on the correct reasoning. We distill knowledge from the reasoning attention to our VQA model through a similarity loss term, that encourages question and reasoning attention alignment. Our model can accurately capture the visual components required to find the correct answer and produce interpretable, human-like attention maps, thus boosting baseline performance. We evaluate our pipeline both quantitatively and qualitatively on the Visual Commonsense Reasoning dataset \cite{vcr}, a large-scale dataset for cognition-level visual understanding. To the best of our knowledge, we are one of the first works to employ implicit attention guidance, free from explicit grounding supervision in a vision-language transformer setting.

\section{RELATED WORK}

The main VQA paradigm is multi-layer transformers operating on joint image-text embeddings \cite{uniter,lxmert,oscar,uniter}. These methods benefit from extensive pre-training on large-scale VL datasets, to extract meaningful image-text representations and align visio-linguistic clues. One notable example is VL-BERT \cite{VLBERT}, a model that is pre-trained on text-only corpora with standard Masked Language Modeling (MLM) as well as visual-linguistic corpora via predicting randomly masked words and Regions of Interest (RoIs) of the image. \par

Despite superior performance, state-of-the-art VQA models can often make decisions by relying on shortcuts and statistical regularities instead of comprehending the scene as demonstrated in \cite{shortcuts}. Similarly, authors in \cite{multimodal} identify that VQA models exploit co-occurrences of words in the question and object segments in the image, which they define as multimodal shortcuts. \par
In an attempt to counter shortcuts and language priors, some methods encourage the model to effectively attend to visual components and infer visual relationships. The authors of \cite{hint} align gradient-based explanations with human attention annotations via a ranking loss to guide the network to focus on the correct image regions. The authors of \cite{human} train an attention auxiliary model with ground truth human-labeled attention maps and consequently apply human-like attention supervision to an attention-based VQA model. Another work in this direction \cite{inter} proposes a method that automatically selects region and object annotations from Visual Genome \cite{genome} that serve as labels for implementing visual grounding as an auxiliary task for VQA. In contrast to these approaches, this work mitigates over-reliance on language priors without requiring annotated attention maps. We train our network instead, to look at the image and attend to meaningful visual evidence through reasoning supervision.

\begin{figure*}
  \centering
  \includegraphics[scale=0.67]{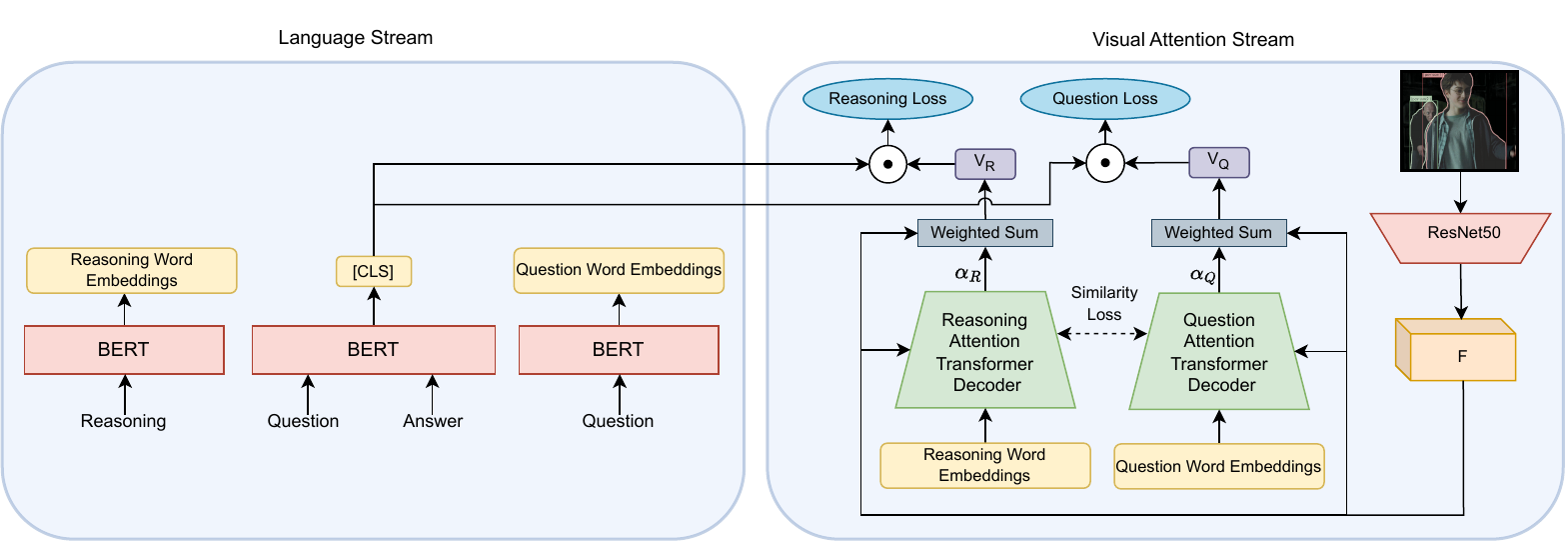}
  \caption{Proposed VAQ architecture: Our model comprises 2 main streams that operate in parallel, a \textit{language} and a \textit{visual} stream. The output of the 2 streams is fused via Hadamard multiplication to obtain the final prediction. During training, we utilize a \textit{reasoning} attention decoder to distil reasoning information into the model, through a similarity loss between question and reasoning-guided attention maps. Reasoning supervision leads in the formation of interpretable attention maps.}
    \label{fig:1}
\end{figure*}

\section{METHODOLOGY}
\label{sec:methodology}

\noindent
\textbf{Problem statement.}
A VQA model is tasked with answering natural language questions from the visual content of a scene. Given a dataset $\mathcal{X}=\{u_i,q_i,a_i,r_i\}_{i=1}^N$ of $N$ images where $u_i \in V$ is the visual input with question $q_i \in Q$, reasoning $r_i \in R$ and groundtruth answer $a_i \in A$, our goal is to learn a function $f: Q \times V \rightarrow \mathbb{R}^A$ that predicts a distribution $P(A)$ over possible answers in $A$. Our proposed pipeline consists of two parallel streams, a \textit{language stream} with model parameters $\theta_L$ and a \textit{visual attention stream} with model parameters $\theta_{V_q}$ and $\theta_{V_r}$ (question and reasoning guided attention decoder that we will discuss next).  During training, we will utilize reasoning supervision as an additional supervisory signal, thus modeling $P(A|u_i,q_i,r_i;(\theta_L,\theta_{V_q},\theta_{V_r}))$ that simplifies to $P(A|u_i,q_i;(\theta_L,\theta_{V_q}))$ at test time. \\

\noindent
\textbf{Language Stream.}
The first stream is language-focused and aims to generate an informative representation of the input question and answer sentence pairs by modeling their relationship. The core of its architecture is a bi-directional 12-layer transformer initialized with weights from BERT~\cite{bert}. It takes a sequence of word embeddings of the question and answer as input (separated by a separation element [SEP]) and adds a sequence positional embedding to each token. The final output feature $x_{[CLS]}$ of the [CLS] element is used to obtain the final pooled linguistic representation. \\

\noindent
\textbf{Visual Attention Stream.}
The visual attention stream consists of two 9-layer transformer decoders. The first one generates an attention vector over the image features guided by the question, and the second an attention vector over the image features guided by the correct reasoning. We take advantage of the cross-attention module to perceive multimodal information and capture relationships between image features and word embeddings. The process is as follows: The image is first processed via the backbone of a ResNet-50-FPN to extract visual appearance features. The output is a feature map $\mathcal{F} \in \mathbb{R}^{H \times W \times 256 }$, which we treat as a sequence of 256-dimensional image features. Following \cite{VLBERT}, a visual geometric embedding is added to each input token to inject 2D awareness into the model. We also encode the question and correct reasoning language tokens via a pre-trained BERT model, which yields a 786-dim representation for each word. 

Question and reasoning word embeddings are used as input to the corresponding question and reasoning transformer decoders (functioning as query tokens). The image visual features $\mathcal{F}$ are used to generate the keys and values. Then, the attention weights are calculated based on the pairwise similarity of the query and key elements. The output of each decoder is an attention distribution over the image regions $\alpha \in \mathbb{R}^{H \times W}$, conditioned on either the word embeddings of the question (referred to as $\alpha^{Q}$) or the word embeddings of the correct reasoning (referred to as $\alpha^{R}$).
In practice, to obtain the final attention vectors $\alpha^{Q}$ and $\alpha^{R}$, we compute the average per-head attention of the last layer generated by the $[CLS]$ token over the image features. 

The generated attention map $\alpha^{Q}$, is then used to take the \textit{weighted sum} over the image features $\mathcal{F}$, which is passed through a linear layer to obtain the final \textit{attended-by-the-question} representation of the image, $V_{q}$. The same operation is performed to obtain the \textit{attended-by-the-reasoning} image representation $V_{r}$. Formally,
\begin{equation}
\begin{split}
   V_{q}=Linear(\alpha^{Q} \odot \mathcal{F}) \\
   V_{r}=Linear(\alpha^{R} \odot \mathcal{F})
\end{split}
\end{equation}\\

\begin{figure*}
  \centering
  \includegraphics[scale=0.27]{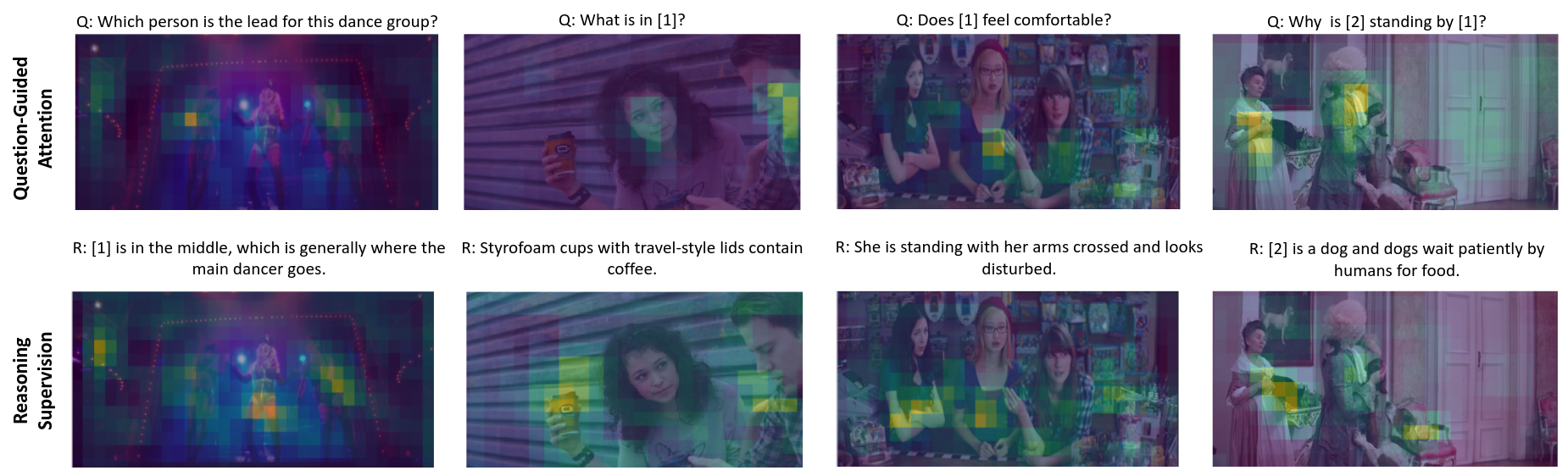}
  \caption{Comparison of question-guided attention maps (only the question attention decoder) before \textit{(first row)}, and after finetuning with reasoning supervision \textit{(second row)}. We observe that the finetuned model is able to attend to informative regions.}
    \label{fig:4}
\end{figure*}

\noindent
\textbf{Combining Language and Visual Streams.}
The model outputs two separate predictions, one conditioned on the question $y_{q}$ and the other on the correct reasoning $y_{r}$. These are produced by fusing the outputs of the \textit{language stream} $x_{[CLS]}$ and the \textit{visual attention stream}, via Hadamard multiplication and then passing them through a softmax classifier $s$, thus $y^{q} = s(x_{[CLS]} \odot V_{q})$ and $y^{r} = s(x_{[CLS]} \odot V_{r})$. At test time,  $y^q$ is used to provide predictions over possible answers. \\

\noindent
\textbf{Training.}
Our training pipeline consists of 2 stages. In the first stage, we train with two cross-entropy loss terms $\mathcal{L}_q$ and $\mathcal{L}_r$ w.r.t the ground truth answer $[a_i]$, or

\begin{equation}
   \mathcal{L}_{stage_1} = -\frac{1}{N}\sum_i^N\log(y_i^q)[a_i]-\frac{1}{N}\sum_i^N\log(y_i^r)[a_i]
\end{equation}

In the second stage, we distill knowledge from the reasoning decoder by aligning the attention distributions conditioned on the question $\alpha^{Q}$ to the attention distributions conditioned on the correct reasoning $\alpha^{R}$. To that end, we freeze the weights of the reasoning attention decoder and only fine-tune the question attention decoder (through $\mathcal{L}_q$) while also utilizing an attention similarity loss, formulated as the forward Kullback-Leibler divergence between attention maps $\alpha^{Q}$ and $\alpha^{R}$, or $D_{KL}( \alpha^{Q} || \alpha^{R})$. The complete $\mathcal{L}_{stage_2}$ loss is:

\begin{equation}
   \mathcal{L}_{stage_2} = -\frac{1}{N}\sum_i^N\log(y_i^q)[a_i]+\frac{1}{N}\sum_i^N \alpha^Q_i \log(\frac{\alpha^Q_i}{\alpha^R_i})
\end{equation}

The whole process is illustrated in Figure \ref{fig:1}. Our model is trained for 11 epochs for stage 1 and then finetuned for 5 more epochs in stage 2. 

\section{EXPERIMENTS}

\noindent
\textbf{Dataset.}
We validate our VQA model on the Visual Commonsense Reasoning dataset \cite{vcr}, which consists of 290k QA problems derived from 110k movie scenes. Four possible answers and four rationales are provided for each question, but we use only the correct rationale/reasoning. Note that reasoning is only used during training as additional supervision. \\

\noindent
\textbf{Quantitative Evaluation.}
Results in terms of model accuracy are reported in Table \ref{table:1}. Our baseline model (only the question decoder) achieves $61.2\%$ accuracy on the validation set. Finetuning by aligning question and reasoning attention distributions yields $63.9\%$, that is a $2.7\%$ absolute improvement, thus demonstrating the benefit of reasoning supervision. We note that our main goal is to propose a novel training strategy for boosting a VQA model's visual explanatory strength by exploiting reasoning as an alternative supervisory signal. Thus, we do not directly compare to methods such as \cite{VLBERT,uniter,merlot} that contain a larger number of parameters, leverage large-scale VL and video pretraining or ground-truth object bounding boxes. For comparison, the best performance reported in R2C \cite{vcr} was $63.8\%$. \par

\begin{table}
 \centering
 \begin{tabular}{l|c}
\toprule
Model & Acc(\%) \\
\midrule
   \textit{Baseline model} & 61.2  \\ 

   \textit{Reasoning Supervision} & $\mathbf{63.9}$ \\ 
\bottomrule
  \end{tabular}

 \caption{Accuracy of the baseline and our proposed model finetuned with \textit{reasoning supervision} on VCR.}
  \label{table:1}
\end{table}

\begin{table}
 \centering
 \begin{tabular}{l|c|c}
\toprule
   Model & Acc(\%) & ($+$\textit{masking}) Acc(\%)   \\
\midrule
   \textit{Baseline model} & 61.2 & 59.3 $(-1.9)$\\ 
    \textit{Reasoning Supervision} & 63.9 & 61.1 $(-2.8)$\\
\bottomrule
  \end{tabular}

 \caption{Performance drop on the VCR validation set due to object masking.} 
    \label{table2}
\end{table}

To further investigate our model's ability to leverage the visual modality, we perform an ablation study where we mask the visual features of the objects/people referenced by the question at test time and measure the effect on accuracy. Results are reported in Table \ref{table2}. We observe that the baseline VQA model (that does not fully alleviate the lack of visual grounding) suffers a lesser performance degradation of $1.9\%$ compared to $2.8\%$ for our finetuned model (on reasoning supervision). This is a different manifestation of the fact, that the baseline model is over-reliant on the language modality, thus performance is penalized less when visual information is not available due to object masking. \\

\noindent
\textbf{Visual Results.}
In Fig. \ref{fig:4}, we visualize attention maps ($\alpha^{Q}$) for both the baseline model \textit{(above)} and finetuned model (with reasoning supervision) \textit{(below)}. The correct reasoning can intuitively provide important guidance during training. For example, for the question \textsc{(Q: Which person is the lead for this dance group?)}, the reasoning \textsc{(R: [1] is in the middle, which is generally where the main dancer goes)}) clearly explains the dynamics of different elements of the scene. This information is distilled to our VQA model through our attention similarity loss. In Fig. \ref{fig:4}, we observe that after fine-tuning, visual attention improves. Our method is able to produce interpretable, human-like attention maps, thus being able to predict the correct answer by perceiving relevant visual concepts.

\section{CONCLUSION}

In this work, we alleviate the lack of visual grounding through reasoning supervision. This additional supervision takes the form of textual justifications of the correct answer and it's already available for VCR datasets. We incorporate a similarity loss that encourages the alignment between the visual attention maps (guided by the question and correct reasoning) thus improving the model's visual perception capability. We demonstrate qualitatively and quantitatively that reasoning information can lead to interpretable attention maps and performance increase for visual question answering.

\nocite{squint}

\bibliographystyle{IEEEbib}
\bibliography{strings,refs}

\end{document}